\documentclass{hooml2022} 

\title[Black Box Lie Group Preconditioners]{Black Box Lie Group Preconditioners for SGD}

\optauthor{%
\Name{Xilin Li} \Email{xilin@meta.com}\\
\addr Facebook SUN 103, 900 Discovery Wy, Sunnyvale, CA 94089}

\begin{document}

\maketitle

\begin{abstract}%
A matrix free and a low rank approximation preconditioner are proposed to accelerate the convergence of stochastic gradient descent (SGD) by exploiting curvature information sampled from Hessian-vector products or finite differences of parameters and gradients similar to the BFGS algorithm.    
Both preconditioners are fitted with an online updating manner minimizing a criterion that is free of line search and robust to stochastic gradient noise, and further constrained to be on certain connected  Lie groups to preserve their corresponding symmetry or invariance, e.g., orientation of coordinates by the connected general linear group with positive determinants.  
The Lie group's equivariance property facilitates preconditioner fitting, and its invariance property saves any need of damping, which is common in second-order optimizers, but difficult to tune. 
The learning rate for parameter updating and step size for preconditioner fitting  are naturally normalized, and their default values work well in most situations.   
\end{abstract}


\section{Introduction}

Second-order optimization for machine learning models with millions of free parameters to learn is challenging. Off-the-shelf convex optimization algorithms \cite{Boyd}, to name a few, quasi-Newton ones like the Broyden–Fletcher–Goldfarb–Shanno (BFGS) and its limited-memory version, LM-BFGS, conjugate gradient (CG) and its nonlinear versions such as Hessian-free (HF) optimization \cite{Martens12}, are successful for small-scale convex mathematical optimization problems, but not commonly used for large-scale stochastic optimization problems like those that arise from machine learning (ML). One of the most prominent hindrances is their dependence on the line search step. The cost functions in many ML models, e.g., variational and reinforcement learning models, are simply defined as expectations, and the only way to evaluate them is via Monte Carlo (MC) sampling averages, which could have large variances. An optimizer relying on line search to ensure convergence could be problematic for them. This issue is less grave in ML problems like classification and regression. Hopefully, line search becomes benign with increased sample sizes. But, large sample sizes bring other new issues, e.g., increased generalization gap and low computational efficiency. Empirical results suggest that the plain SGD is a highly efficient optimizer for most ML problems. Still, conceivably, SGD will converge slowly for problems with large eigenvalue spread once the solution is located in a basin of attraction. Regret optimizers like RMSProp and Adam \cite{adam} converge faster but are empirically shown to generalize less well on many problems. Reducing the generalization gap between SGD and Adam is still an active topic \cite{Adabelief}, although not the focus here. We believe that all SGD needs is a good preconditioner to accelerate its convergence around the basin of attraction, without undermining its generalization capacity. The curvature information for preconditioner fitting can be sampled from the Hessian-vector products or finite differences of parameters and gradients similar to the BFGS algorithm. As discussed above, we may not be able to construct a preconditioner in a deterministic way as in BFGS since line search could be problematic.
Thus, we adopt a more general and gradient noise robust preconditioner fitting criterion proposed in \cite{Li18},  and fit the preconditioner online with another `gradient descent' algorithm.   
The point is that we should not turn the preconditioner fitting problem into a more difficult and computationally expensive one than the original parameter learning problem. Here, the Lie group is the perfect tool for preconditioner fitting. Note that the `gradient descent' on a Lie group is similar to but different from the common gradient descent in Euclidean space. It is achieved by applying a series of small transforms via multiplication with $I +\mu G$, where $\mu$ is a small scalar, and $G$ is the group generator.  
A Lie group is actually a quite friendly object to work with. Moving a preconditioner around any point on a Lie group just behaves as moving it around the identity element of the group, i.e., the identity matrix $I$.      
This is known as  the equivariance property of a Lie group. 

\section{Background}

\subsection{The Notations}

We are to minimize a loss function defined via expectation as 
$
f(\theta) =  E_z[\ell(\theta, z)] 
$, where $\theta\in \mathbb{R}^n$ is the parameter vector to be optimized, and $z$ is a random vector that can be sampled to evaluate the loss $\ell(\theta, z)$. 
We always assume that the considered problem is second-order differentiable. 
To simplify the handwritings, we just use $\hat{f}(\theta)$ to denote one sampled noisy evaluation of $f(\theta)$. 
Then, one step of SGD with learning rate $\mu$ and an optional positive definite preconditioner $P$ writes as
\begin{equation}
\theta_{i+1} = \theta_{i} - \mu P \,{\partial \hat{f}(\theta)}/{\partial \theta}\left.\right|_{\theta=\theta_i}
\end{equation}
where $i$ is the iteration index,  $\mu>0$ is the learning rate, and $P$ typically is a variable or adaptive preconditioner. 
Once the solution enters a basin of attraction centered at a local minimum $\theta^*$, we can approximate the iteration step in (1) as   
\begin{equation}
\theta_{i+1} - \theta^* \approx (I - \mu P \hat{H}) ( \theta_{i} - \theta^*)
\end{equation}
where $\hat{H}=\frac{\partial^2 \hat{f}(\theta)}{\partial \theta^T \partial \theta}\left.\right|_{\theta=\theta^*}$ is the sampled Hessian at the local minimum.
Conceivably, the eigenvalue spread of $ P \hat{H}$ largely determines the speed of convergence of the quasi-linear system in (2). 
Nearly quadratic convergence is possible if we can figure out a good approximator for  ${H}^{-1}$. 
But, $\hat{H}$ is a noisy Hessian, and not necessarily positive definite, even if the exact one at  $\theta^*$, i.e., $H$, is. 

\subsection{The Preconditioner Fitting Criterion}

We adopt the preconditioner fitting criterion proposed in \cite{Li18}. Following their notations, let $\delta g$ be the perturbation of gradient associated with parameter perturbation $\delta \theta$. Then, this fitting criterion is 
\begin{equation}
c(P) = E_{\delta\theta}[\delta g^T P \delta g + \delta \theta^T P^{-1} \delta \theta]
\end{equation}
With auto differentiation tools, we simply replace pair $(\delta\theta, \delta g)$ with $(v, \hat{H}v)$, where $v$ is a random vector, and $\hat{H}v$ is the noisy Hessian-vector product, which can be evaluated as computationally cheap as the gradients. 
Criterion (3) only has one positive definite solution, $P = (H^2 + E_v[\epsilon^2])^{-1/2}$, even for indefinite $H$, where $\epsilon = \hat{H} - H$ is a stochastic noise term.  
Hence, this preconditioner automatically damps gradient noise. 
It is worth noting  that criterion (3) gives the same preconditioner used in   equilibrated SGD (ESGD) \cite{Dauphin15} and AdaHessian \cite{adahessian} when $P$ is diagonal, i.e., 
$E[v\odot v]\oslash E[(\hat{H}v)\odot (\hat{H}v)]$, where $\odot $ and $\oslash$ denote element-wise product and division, respectively. 

\subsection{Preconditioners on Lie Groups}

It is natural to fit the preconditioner on Lie group for several reasons. 
First, let us rewrite (1) as 
$
 P^{-1/2}\theta_{i+1} = P^{-1/2}\theta_{i} - \mu  \,{\partial \hat{f}(\theta)}/{\partial (P^{-1/2}\theta)}\left.\right|_{\theta=\theta_i}
$. 
Now, it is clear that a preconditioned SGD equals SGD with a new set of  coordinates defined by $\vartheta = P^{-1/2}\theta$. 
A coordinate change consists of rotations and scalings, i.e., operations on the orthogonal group $O(n)$ and the group of nonsingular diagonal matrices. 
Let us represent this coordinate transform with matrix $Q^{-1}$, and accordingly, $P=Q^TQ$. 
Now, we can pursue  a variable $Q$ on the Lie group to fit this coordinate transform. 

Second, preconditioned SGD also can be viewed as SGD with transformed features when the parameters to be learned are a list of affine transform matrices \cite{psgd_affine}. 
Specifically, the most commonly used feature transformations, e.g., whitening, normalization, and scaling, can be represented as matrices on Lie groups.  
For example, the popular batch normalization operation \cite{batchnorm} can be represented as a sparse preconditioning matrix on a Lie group where only the diagonal and last column can have nonzero values \cite{psgd_affine}. Again, the Lie group arises as a natural object to work with. 

Lastly, the Lie groups have two wonderful properties suitable for our task. As with any group, a certain Lie group preserves certain symmetry or invariance. For example, with $Q\in GL^+(n, \mathbb{R})$, the general linear group with positive determinant,  $\vartheta$ and $\theta$ will always have the same orientation. This saves the need for any damping or similar remedies for avoiding degenerated solutions since $P=Q^TQ$ is guaranteed to be always invertible. The equivariance property of Lie groups further facilitates the preconditioner fitting. The same group generator, i.e., the one at the identity matrix, can be used to move a preconditioner on any point of the Lie group.

The preconditioners proposed in \cite{psgd_affine} can only be applied to a list of affine transform matrix parameters. Although many machine learning models indeed exclusively consist of affine transforms and nonlinear activation functions, this is not always the case. Also, practically, it is inconvenient to reparameterize many existing modules, e.g., a convolutional layer, into their equivalent affine transform forms. Hence, two types of novel black box preconditioners are proposed in this paper.
  
\section{Black Box Lie Group Preconditioners}

\subsection{Simple Matrix Free Preconditioners}

The term `matrix free' suggests that we do not explicitly form the matrix representations. 
The Lie groups keep to be  abstract forms, i.e.,   transformations in  vector space,  $T: \mathbb{R}^n\mapsto \mathbb{R}^n$. 
The following theorem gives one systematic way to construct such sparse Lie group preconditioners. 

\emph{Proposition 1:} Let $K=\{\sigma_1, \ldots, \sigma_m\}$ be a subgroup of the permutation group $S_n$. 
Then, linear transform
$
T: \mathbb{R}^n \mapsto \mathbb{R}^n, \quad T(x|a_1, \ldots, a_m) = \sum_{i=1}^m a_i\odot \sigma_i(x)
$, forms a subgroup of $GL(n, \mathbb{R})$ parameterized with $\{a_1, \ldots, a_m\}$ if $T(\cdot|a_1, \ldots, a_m) $ is bijective, where both $a_i$ and $x$ are in $\mathbb{R}^n$.

Proposition 1 is proved by showing that   $T$ can be reduced into the direct sum of  $\lceil n/|K|\rceil$ smaller irreducible Lie groups, where $|K|$ is the order of  $K$. 
We list a few prominent examples below. 

\emph{Example 1}: the group of invertible diagonal matrices. We must have $K=\{e\}$ if $|K|=1$, where  $e$ is the identity element of $S_n$, i.e., $e(x)=x$. 
Then, $T$ simply has a diagonal matrix representation, i.e., $T(x|a_1) = {\rm diag}(a_1) x$. 
Criterion (3) gives the preconditioner in  ESGD \cite{Dauphin15} and AdaHessian \cite{adahessian} as a special case when $P$ is on this group. 

\emph{Example 2}: the group of X-shape matrices. 
Let $K=\{e, \sigma_f\}$, where $\sigma_f$ denotes the flipping permutation. 
Then, we can show that 
\begin{align*}
T(\cdot|a, b)  T(\cdot|u, v) & = T(\cdot|a\odot u + b \odot \sigma_f(v), a\odot v + b \odot \sigma_f(u)) \\
T^{-1}(\cdot|a, b) & = T(\cdot|  {\sigma_f(a)\oslash c , -b\oslash c})
\end{align*}
where $c = a\odot \sigma_f(a) - b\odot \sigma_f(b)$. 
Clearly, such transforms form a Lie group if invertible, i.e., no element of $c$ is zero. 
The matrix representation of this $T$ only has nonzero diagonal and anti-diagonal elements, thus the name X-shape matrix.

\emph{Example 3}: the butterfly matrix \cite{butterfly}. For an even $n$, subgroup $K=\{ e, s_{n/2} \}$ induces  a Lie group whose representations are invertible butterfly matrices, where $s_{n/2}$ denotes circular shifting by $n/2$ positions.  This group of matrices are the building blocks of the Kaleidoscope matrices \cite{butterfly}. 

\emph{Example 4}: the plain dense invertible matrix. The group $GL(n, \mathbb{R})$ can be recovered by letting  $K=\{ e, s_{1}, \ldots, s_{n-1} \}$, where  $s_{i}$ denotes circular shifting by $i$ positions.  

The group $GL(n, \mathbb{R})$ is too expensive for large-scale problems. The group of diagonal matrices, also called the Jacobi preconditioner in its matrix form, is sparse enough, but empirically shown to be less effective without the help of momentum for certain machine learning problems. We are mostly interested in the cases with $2\le |K|\le 4$. These Lie groups are sparse enough, yet simple enough to derive their inverse explicitly, and at the same time could significantly accelerate the convergence of SGD by shortcutting gradients separated far away in positions.

\subsection{Low Rank Approximation Preconditioner}

Low-rank approximation (LRA) is a standard technique in processing large-scale matrices. Commonly adopted forms of positive definite low-rank approximation, e.g., $P=\rho I + UU^T$, cannot always be factorized as $P=Q^TQ$ such that $Q$ is on certain Lie groups, where $\rho>0$ is a small positive number. Furthermore, this is not an effective form of approximation for reducing eigenvalue spread. The Hessians in many real-world problems typically have a few very large and very small eigenvalues, i.e., tails on both ends of the spectra. But, all the eigenvalues of $P$ in this form are lower bounded by $\rho$. Thus, it can only fit one tail of the spectra when rank$(U)\ll n$.

For this reason, we propose a new low-rank approximation with form $Q= \rho (I + UV^T)$, where $\rho$ is not necessarily small nor positive, and $U$ and $V$ have $r$ columns with $r\ll n$. To justify this form of approximation, we need to show two facts. First, preconditioner $P=Q^TQ$ with this form can fit both tails of the spectra of Hessian. Second, we can update this preconditioner on Lie groups.

\emph{Proposition 2:} Preconditioner $P=Q^TQ$ with $Q =\rho( I+UV^T)$ can have positive eigenvalues arbitrarily larger than $|\rho|$ and arbitrarily smaller than $|\rho|$ with proper $U$ and $V$. 

\emph{Proposition 3:}  If $\rho\ne 0$ and $(I+V^TU)^{-1}$ or $(I+U^TV)^{-1}$ exists, $A_V(d, U)=\rho(I+UV^T)$ defines a subgroup of $GL(n, \mathbb{R})$ parameterized with $\rho$ and $U$. Similarly, $A_U(\rho,V)=\rho(I+UV^T)$ defines another  subgroup of  $GL(n, \mathbb{R})$ parameterized with $\rho$ and $V$. 

Proposition 2 can be shown to be true just by checking the simplest case with $n=2$ and $r=1$. 
Proposition 3 can be shown to be true by verifying the four requirements of being a Lie group. Notably, the Lie algebra  is closed, e.g., 
$
<U_1V^T, U_2V^T >  =  (U_1V^TU_2 - U_2V^TU_1)V^T
$ on Lie group $A_V(U)$, where $<\cdot>$ is the Lie bracket. 
The condition that $(I+V^TU)^{-1}$ or $(I+U^TV)^{-1}$ exists is  to ensure that $I+UV^T$ is invertible as shown by the Woodbury matrix identity. 
The rotation ambiguity, i.e., $I+UV^T=I+UR(VR)^T$ for any $RR^T=I$, can be removed by assuming $I+V^TU$ having the Schur decomposition form, if necessary.

\section{Practical Considerations}

All the above proposed preconditioners can be fitted online by minimizing criterion (3) with `gradient descent' on the Lie groups. 
Different from the common gradient descent, moving an object on Lie group is achieved by multiplying it with $I + \mu G$, where  $G$ is the group generator, and $\mu$ is small enough such that $\|\mu G\|<1$. Since a series of such small movements trace a curve on the Lie group manifold, we see that $G$ indeed  always is in the tangent space of the group as Lie algebra is closed. 
We do not  detail the math derivations here.
They are   documented in our supplemental materials.  

We would like to point out that damping is neither necessary nor generally feasible on any Lie group, although it is widely used in  other second-order optimizers to avoid degenerated solutions. 
On one hand, by fitting $Q$ on a connected Lie group, $P=Q^TQ$ cannot be singular. 
On the other hand, damping could be  incompatible with certain forms of Lie groups. 
We may not always be able to find another $Q'$ on the same group such that $Q'^TQ' = Q^TQ + \lambda I$ with $\lambda>0$.
Actually, criterion (3)  damps the gradient noise naturally \cite{Li18}.
This spares us the trouble on tuning a damping schedule. 

Yet, gradient clipping could be helpful to encourage stability.  
The quadratic approximation leading to the quasi-linear system  (2) can only be valid within a certain region around $\theta$. 
Thus, $\| \delta\theta \| = \mu \| P {\partial \hat{f}(\theta)}/{\partial \theta}\| $ should be small enough such that $\theta + \delta\theta$ still locates in this trust region. 
We can either adjust $\mu$, or equivalently, clipping $\|P {\partial \hat{f}(\theta)}/{\partial \theta}\|$, to ensure that $\| \delta\theta \|$ is small enough. 

Lastly, both the learning rate for parameter updating and step size for preconditioner fitting are naturally normalized to be in range $(0, 1)$. 
Step size $0.01$ always is a good initial guess. 
It is not difficult to develop enough experiences for the user on setting them  given an optimization problem. 

\begin{table}
  \begin{center}
    \caption{Test classification error rates (\%) on  MNIST with LeNet5  over ten runs. Lower is better. }
    
    \label{tab:table1}
    \begin{tabular}{c|c|c|c|c|c} 
      SGD+M & Adam \cite{adam}  & KFAC \cite{kfac} & PSGD \cite{psgd_affine} & PSGD XMat & PSGD LRA   \\
       $0.96\pm 0.07$ & $0.88\pm 0.07$ & $0.84\pm 0.07$ & $0.74\pm 0.06$ & $0.78\pm 0.06 $  & $0.78\pm 0.07$ 
    \end{tabular}
\end{center}
\end{table}

\begin{table}
  \begin{center}
    \caption{Test classification accuracy  (\%) on CIFAR10 with ResNet18 over $16$ runs. We test stage and cosine learning rate schedulers, and removal of shortcut connections.  Higher is better. }
    
    \label{tab:table1}
    \begin{tabular}{c|c|c|c|c|c|c} 
     & with shortcut, stage lr &  with shortcut, cos lr & without shortcut, cos lr     \\
    SGD+M  & $95.07\pm 0.12$ & $95.51\pm 0.11$ & $94.97\pm 0.16$ \\
    PSGD LRA & $95.43\pm 0.12$ & $95.54\pm 0.10$ &  $ 95.36\pm 0.09$\\
\end{tabular}
\end{center}
\end{table}

\section{Experimental Results}

We have tested the proposed preconditioners on four tasks: 1) the MNIST handwriting digit recognition with LeNet5; 2)  the CIFAR10 image classification with ResNet18; 3) a large scale logistic regression problem, and 4) the delayed XOR problems \cite{lstm} with sequence length $64$. 
Detailed experimental setups and results are put in the appendices. 
Tables 1 and 2 summarize the image classification results. 
Preconditioned SGD (PSGD) also outperforms LM-BFGS on the third problem.    
The last problem is well known to be  difficult for either simple or gated recurrent neural networks (RNNs). 
It seems that only PSGD can reliably solve this type of pathological problem. 
These initial results are encouraging for second-order  optimizations using preconditioners on the Lie groups.

\section*{Acknowledgments: }

I would like to thank Yaroslav Bulatov for his encouragements of the work on preconditioned SGD, and Omead Pooladzandi for insightful discussions. 
 
\bibliography{psgd}

\newpage 

\section*{Appendix A: More Experimental Results}

\subsection*{A.1: MNIST Handwriting Digit Recognition}

To have a fair comparison between the diagonal and low rank approximation (LRA) preconditioners, we slightly upgrade  $Q$ in the LRA preconditioner to form \[ Q = {\rm diag}(d) (I + UV^T) \] where $d$ is a vector. This form of $Q$ cannot form a Lie group. Still, its two factors, ${\rm diag}(d)$ and $I+UV^T$, can be fit on their Lie groups. Now, the diagonal preconditioner is a special case of this LRA one with order $r=0$. We have tested orders $r=0,1,2,5,$ and $10$. The batch size is $64$. Totally ten epochs of training are performed. The learning rate for parameter updating is annealed exponentially from $0.1$ for the first epoch to $0.001$ for the tenth epoch. The step size for preconditioner fitting is annealed exponentially from $0.1$ for the first epoch to $0.01$ for the tenth epoch. Preconditioned gradient norm is clipped to $10$ if too large. No momentum is used. Figure 1 summarizes the test classification error rates for preconditioners with different orders of approximations over $50$ runs. From Fig.~1, we see that the simple diagonal preconditioner performs well. Still, LRA brings marginal gains up to order $r=5$. This cost function is fairly `flat' since the only nonlinearities in LeNet5 are two piece-wise linear functions, i.e., the activation function ReLU and max pooling one. Only the cross entropy loss introduces the `curvature'.

\begin{figure}
\centering
\includegraphics[width=0.5\textwidth]{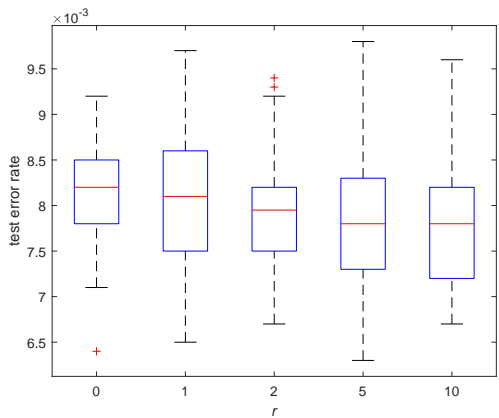}
\caption{MNIST test classification error rates over $50$ runs using preconditioners with different orders of LRA. The one with  order $0$ reduces to the diagonal preconditioner. 
Table 1 reports results of the one with $r=5$. Lower is better.    }
\end{figure}

\subsection*{A.2: CIFAR10 Image Classification with ResNet18}

We follow the implementations of the Adabelief algorithm \cite{Adabelief}  to test preconditioned SGD (PSGD) on the CIFAR10 image classification task with the ResNet18 model. 
Training code of Adabelief is available at {\small \url{https://github.com/juntang-zhuang/Adabelief-Optimizer}}. 
One main difference from the implementations in \cite{Adabelief} is that we reduce the learning rate by tenfold twice for all the optimizers, while the original stage learning rate scheduler only anneals the step size once.
We also consider the cosine learning rate scheduler, which helps SGD to achieve the state-of-the-art (SOTA) test accuracy of about $95.5\%$. 
Training and testing accuracy convergence curves over $16$ runs are plotted in Fig.~2. 
We only show the results of PSGD and SGD here as SGD is known to achieve the SOTA results for this problem. 

For PSGD, we use step size $0.02$ for parameter updating and $0.01$ for preconditioner fitting.    
The preconditioner is only updated once per ten iterations, and thus its overhead over SGD is marginal.    
The same momentum factor, $0.9$, is used for both SGD and  PSGD.
Since the step size in PSGD is normalized, we update the momentum as $m\leftarrow 0.9m + 0.1 g$, instead of $m\leftarrow 0.9m +  g$ as in the SGD. 
No gradient clipping is used. 
Weight decay is realized by adding the L2 regularization term $0.5 \lambda \theta^T\theta$ to the cross entropy loss. 
We have found that $\lambda$ between $  0.01$ and $  0.02$ performs the best. 

From Fig.~2, we observe that 
SGD performs very well with the cosine learning rate scheduler.  
This is expected as these residual networks are highly evolved to be first-order optimizer friendly.  
The extensive use of piece-wise linear functions, residual connections, and batch normalizations make these models fairly `flat' and resemble shallow models, instead of deep ones. 
Still, PSGD slightly outperforms SGD  when we remove the shortcut connections or use a less-tuned learning rate scheduler, e.g., the stage one here.


\begin{figure}
\centering
\includegraphics[width=0.63\textwidth]{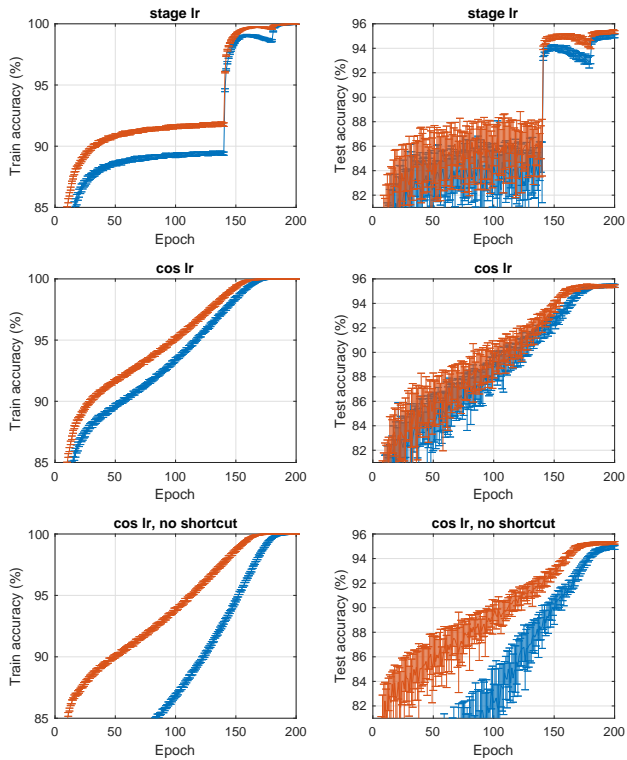}
\caption{CIFAR10 image classification with ResNet18.  
The order of low rank Hessian approximation  is $10$. 
Mean and variance are estimated over $16$ runs. 
Higher is better. }
\end{figure}

\subsection*{A.3: A Large Scale Logistic Regression Problem}

\begin{figure}
\centering
\includegraphics[width=0.55\textwidth]{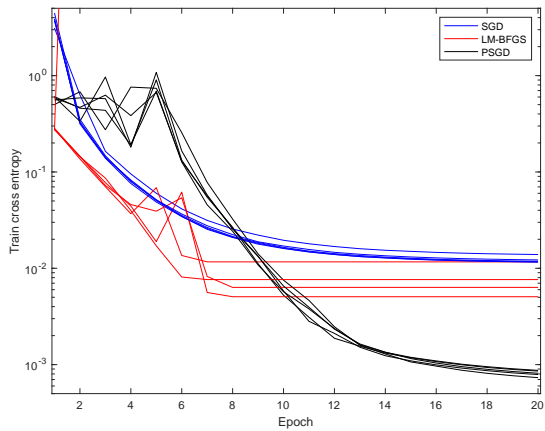}
\caption{Typical convergence curves on the logistic regression problem.  Lower is better.  
When comparing the convergence speed, one should be aware that  one step of LM-BFGS may take up to ten iterations, while SGD and PSGD always one iteration per step. 
}
\end{figure}

We use logistic regression to solve the MNIST image classification problem. 
Let $x$ be the vector of image with length $28^2$.
Instead of regression on vector $x$, we do the regression on the outer product vector of $x$, which has length $28^4$. 
This significantly increases the test classification accuracy, but leads to a large regression matrix with over six million coefficients. 

We compare PSGD with the algorithm of choice for this type of problem, LM-BFGS. 
No momentum is considered since this is the case for LM-BFGS. 
The train batch size is $500$. 
It is tricky to select the initial learning rate for LM-BFGS even we exponentially anneal down it. 
We have found that LM-BFGS diverges on roughly one third of the trials with initial learning rate $0.1$, but $0.05$ is too small and may lead to worse performance than SGD. 
For PSGD, we consider the LRA preconditioner with order $10$, and set the learning rates for parameters and preconditioner to $0.05$ and $0.1$, respectively. 
Since LM-BFGS might diverge with learning rate $0.1$, we only show a few typical convergence curves of SGD, LM-BFGS and PSGD in Fig. 3.
LM-BFGS converges to regression losses a few times smaller than that of SGD. 
PSGD could converges to  losses about one order of magnitude lower than that of SGD and LM-BFGS. 
Regarding test classification error rate, we have $2.37\%\pm 0.08$,  $2.09\%\pm 0.18$, $1.98\%\pm 0.08$ for SGD, LM-BFGS and PSGD, respectively, averaged over ten runs. 
Again, LM-BFGS outperforms SGD, and PSGD performs the best on the test classification error rate as well.  

\subsection*{A.4: The Delayed XOR Problem}

\begin{figure}
\centering
\includegraphics[width=0.5\textwidth]{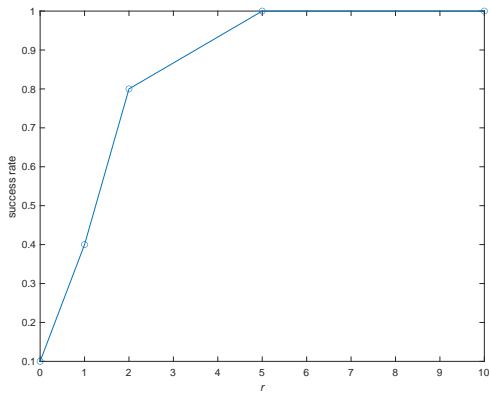}
\caption{Success rate over ten runs in solving the XOR problem with a simple RNN and LRA preconditioners of orders $0,1,2,5$ and $10$.  Higher is better.  }
\end{figure}

We consider the delayed XOR problem proposed in the LSTM paper \cite{lstm}. 
The task is to predict the XOR relation of $a$ and $b$ scattered randomly and far away in a long sequence.  
It is a tough problem because it cannot be `partially solved' since memorizing either $a$  or $b$ alone does not help to predict XOR$(a, b)$. 
This problem is equally challenging to both simple and gated recurrent neural networks (RNNs). 

A simple RNN with $30$ hidden states is adopted. 
The sequence length is $64$. 
Tested orders of low rank approximation are  $r=0,1,2,5,$ and $10$. 
Both step sizes for parameter and preconditioner updating are fixed at $0.01$. 
The gradient clipping threshold is set to $1$. 
No momentum is used. 
A trial fails if it does not converge within $100,000$ iterations.  
The success rates over ten runs for each $r$ are plotted in Fig.~4. 
This example shows a typical problem where the diagonal preconditioner struggles, while a low-order Hessian approximation works perfectly for preconditioning. 


\section*{ Supplementary Materials:}

Derivations of the two preconditioners are put at 
\url{https://drive.google.com/file/d/1CTNx1q67_py87jn-0OI-vSLcsM1K7VsM/view?usp=sharing}.

\end{document}